\documentclass{article}

\usepackage[preprint]{neurips_2026}

\usepackage[utf8]{inputenc}
\usepackage[T1]{fontenc}
\usepackage{hyperref}
\usepackage{url}
\usepackage{booktabs}
\usepackage{amsmath}
\usepackage{amsfonts}
\usepackage{amssymb}
\usepackage{amsthm}
\usepackage{nicefrac}
\usepackage{microtype}
\usepackage{xcolor}
\usepackage{graphicx}
\usepackage{subcaption}
\usepackage{multirow}
\usepackage{placeins}

\hypersetup{hidelinks}

\theoremstyle{definition}
\newtheorem{lemma}{Lemma}

\newcommand{\rms}{\mathrm{rms}}
\newcommand{\rad}{\mathrm{rad}}

\title{Dense Supervision Is Not Enough: The Readout Blind Spot in Looped Language Models}

\author{%
  Rituraj Sharma \\
  Virginia Tech \\
  \texttt{rituraj@vt.edu} \\
  \And
  Tu Vu \\
  Virginia Tech \\
  \texttt{tuvu@vt.edu} \\
}

\begin{document}
\maketitle

\begin{abstract}
Looped language models turn hidden states into runtime state: each state is decoded for prediction and fed back into future computation.
This creates a basic supervision question: which state variables does cross-entropy actually control?
We show that dense per-loop cross-entropy controls the variables exposed by the readout, not every variable active in the recurrent transition.
Hidden-state scale gives a concrete failure mode.
Scale-invariant readouts such as RMSNorm and LayerNorm hide radial scale from the immediate cross-entropy loss, while pre-norm residual recurrence continues to carry and update that same scale.
Thus per-loop loss can make early exits usable without controlling recurrent scale.
In 44M and 129M looped transformers without inter-loop normalization, per-loop cross-entropy through RMSNorm readouts still drives final hidden-state norms into the thousands or tens of thousands.
Scale-visible readouts and explicit norm penalties keep norms in the tens, and scale-removing recurrence is the complementary architectural fix.
The resulting design rule is simple: dense supervision trains exits; recurrent scale control requires either making scale visible to a loss or removing it from the loop.
Consistent with this rule, scale-controlled variants achieve lower perplexity at matched inference-depth operating points in our variable-depth benchmarks.
\end{abstract}

% ============================================================================
\section{Introduction}
\label{sec:intro}

Looped language models reuse the same block across recurrent depth.\footnote{Code: \url{https://github.com/Rituraj003/readout-blind-spot}.}
This makes test-time depth a natural compute knob: stop after a few loops when the prediction is easy, or continue iterating when additional computation helps.
Universal Transformers \citep{dehghani2019universal}, Huginn \citep{geiping2025huginn}, Ouro \citep{zhu2025ouro}, LoopFormer \citep{jeddi2026loopformer}, and Elastic Looped Transformer \citep{goyal2026elt} all explore versions of this idea.

Looping changes the role of a hidden state.
In a fixed-depth transformer, a late hidden state is mostly an interface to the output head.
In a looped model, the same hidden state is also runtime state: it is decoded for prediction and then reused as input to future computation.
This makes a simple question unavoidable:
\emph{which parts of the recurrent state does the supervised loss actually control?}

Dense supervision is the natural answer.
Apply cross-entropy (CE) at every loop, and each intermediate state receives direct prediction supervision.
This indeed trains the prediction interfaces: early exits become usable.
But dense CE does not automatically control every variable carried by the recurrent state.
It controls what the readout makes visible.
If an active recurrent variable is hidden by every supervised readout, the model can be trained at every loop while that variable remains underconstrained.

We study this readout blind spot for hidden-state scale.
In pre-norm residual loops, scale is \emph{active}: the residual skip carries it forward and the learned update can change it.
With an RMSNorm or LayerNorm readout, scale is \emph{locally invisible}: multiplying the hidden state by a positive scalar leaves the readout unchanged up to the normalizer epsilon, so the immediate CE loss has essentially no radial derivative.
The result is a visibility--activity mismatch.
In our looped transformers, per-loop CE through normalized readouts still allows hidden-state norms to drift by orders of magnitude when scale remains active in the recurrent path.
The model is supervised at every loop, yet every supervised interface hides the variable that is drifting.

This yields the central design rule of the paper:
\begin{quote}
    \emph{Per-loop cross-entropy trains exit usability; recurrent scale control requires either making hidden-state scale visible to a loss or removing it from the recurrent path.}
\end{quote}

\begin{figure}[t]
\centering
\includegraphics[width=\linewidth]{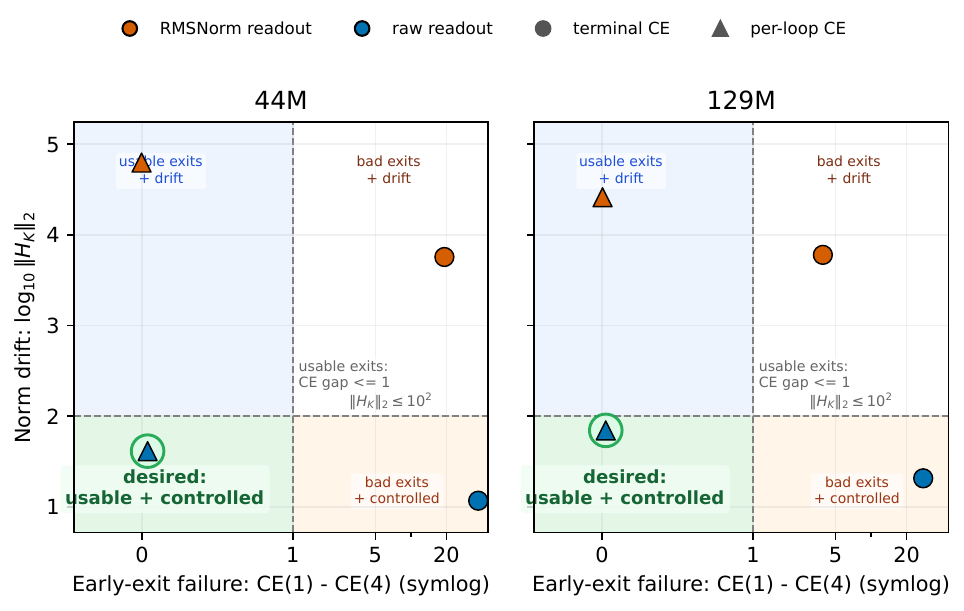}
\caption{\textbf{Exit training and scale control are orthogonal in the core 2$\times$2.}
The x-axis measures early-exit failure, the cross-entropy gap between the first and fourth recurrent loops, $\mathrm{CE}(K{=}1)-\mathrm{CE}(K{=}4)$, where $K$ is the inference loop count; the axis uses a symlog scale.
The y-axis measures final-loop norm drift, with $H_K$ denoting the hidden state after $K$ loops.
Lower is better on both axes.
Per-loop CE moves models left by making intermediate exits usable, while scale-visible readouts move models down by controlling recurrent scale.
The desired region requires both interventions; later norm-penalty and final-only-normalization controls show that the effect is not merely a raw-readout artifact.}
\label{fig:decomp}
\end{figure}

We support this rule with a compact mechanism and controlled ablations.
The mechanism has two components.
Scale-invariant readouts such as RMSNorm or LayerNorm produce zero immediate radial cross-entropy signal, while pre-norm residual recurrence continues to preserve and update scale.
Future-loop losses can still carry indirect radial information, but that signal weakens at large hidden-state norms.
The experiments then test the interventions suggested by this mechanism: exposing scale through raw readouts or explicit norm penalties, or removing scale through normalization inside the recurrent path.

Our contributions are:
\begin{enumerate}
    \item We introduce the readout blind spot as a visibility--activity mismatch: dense CE controls variables exposed by the readout, not every variable active in recurrence.
    In 44M- and 129M-parameter looped language models where hidden-state scale remains active in the recurrent path, per-loop cross-entropy through normalized readouts still produces large norm drift.
    
    \item We instantiate the blind spot for hidden-state scale. Normalized readouts hide scale from the immediate cross-entropy loss, while pre-norm residual recurrence continues to propagate it. Raw readouts, norm penalties, final-only normalization, gradient diagnostics, and recurrent-scale clamps all support this interpretation.
    
    \item We show that scale control improves the variable-depth perplexity--compute frontier. Scale-controlled per-loop models maintain usable exits while achieving lower perplexity than the $K$-invariant RMSNorm baseline at matched inference-depth operating points in our benchmarks.
\end{enumerate}

% ============================================================================
\section{Mechanism: Visible Losses Control Visible Coordinates}
\label{sec:mechanism}

The failure mode is a mismatch between what the recurrent state carries and what the loss can see.
At loop $k$, a hidden state $H_k$ is both a prediction interface and the input to future computation:
\begin{equation}
    H_{k+1}=F_\theta(H_k),\qquad z_k=W_{\mathrm{out}}r(H_k),\qquad
    \mathcal{L}_k=\mathrm{CE}(z_k,y).
\end{equation}
Here $F_\theta$ is the recurrent transition, $r$ is the readout, $W_{\mathrm{out}}$ maps readout states to logits $z_k$, $y$ is the next-token target, and $\mathcal{L}_k$ is the loop-$k$ cross-entropy loss.
A degree of freedom is \emph{immediately visible} if the local readout loss has a derivative in that direction, and \emph{active} if the transition updates it and carries it forward.
Per-loop CE makes each loop predictive, but it does not make every active degree of freedom visible.

\begin{figure}[t]
\centering
\includegraphics[width=\linewidth]{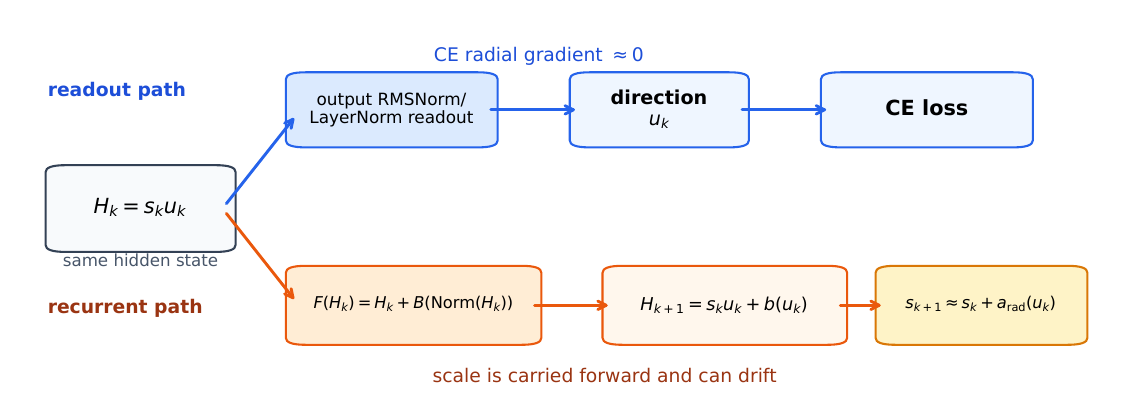}
\caption{\textbf{Visibility--activity mismatch.}
The same hidden state $H_k=s_ku_k$ serves as both a prediction interface and the input to the next recurrent step.
Along the readout path, output normalization removes scale, so the immediate CE loss has approximately zero radial gradient.
Along the recurrent path, the pre-norm residual update still carries the skip state $H_k$, so scale remains active and can drift through the recurrence.
Here $\mathrm{Norm}$ denotes the pre-sublayer normalizer and $B$ denotes the residual branch applied to the normalized state.}
\label{fig:mechanism}
\end{figure}

We study the scale direction.
Let $H=su$, where $s=\|H\|_{\rms}$, $\|u\|_{\rms}=1$, and $\|v\|_{\rms}^2=d^{-1}\|v\|_2^2$ for hidden-vector dimension $d$.
A raw readout preserves scale, $r_{\mathrm{raw}}(su)=su$.
A normalized readout removes it, $r_{\mathrm{norm}}(su)=u$ up to the usual normalization epsilon.
Figure~\ref{fig:mechanism} sketches the two paths the same state takes.

\begin{lemma}[Visibility: scale-invariant readouts remove the immediate radial CE signal]
\label{lemma:radial}
If $r(\alpha H)=r(H)$ for all $\alpha>0$ and $\mathcal{L}(H)=\mathrm{CE}(W_{\mathrm{out}}r(H),y)$ is differentiable at $H\neq0$, then
\begin{equation}
    \langle \nabla_H \mathcal{L}(H),H\rangle=0.
\end{equation}
\end{lemma}
\noindent
The proof is the chain rule applied to $\mathcal{L}(e^tH)$, which is constant in $t$.
With practical LayerNorm/RMSNorm epsilons the identity is approximate, but the radial derivative is small at large hidden norm.

\begin{lemma}[Activity: pre-norm residual loops carry scale]
\label{lemma:expansion}
Let $H=su$ with $\|u\|_{\rms}=1$.
For the epsilon-free pre-norm residual update, with $\mathrm{Norm}$ the pre-sublayer normalizer and $B$ the residual branch,
\[
    F(H)=H+B(\mathrm{Norm}(H)),
    \qquad
    b(u)=B(\mathrm{Norm}(u)),
\]
we have $F(su)=su+b(u)$.
Decompose the residual update into radial and angular parts:
\[
    b(u)=a_{\rad}(u)u+b_\perp(u),
    \qquad
    a_{\rad}(u)=\frac{\langle u,b(u)\rangle}{d},
    \qquad
    \langle u,b_\perp(u)\rangle=0.
\]
If $\|b(u)\|_{\rms}=O(1)$ as $s\to\infty$, then
\begin{align*}
    \text{\emph{scale update:}}\qquad
    \|F(su)\|_{\rms}
    &= s+a_{\rad}(u)+O(s^{-1}), \\
    \text{\emph{direction update:}}\qquad
    \frac{F(su)}{\|F(su)\|_{\rms}}
    &= u+\frac{b_\perp(u)}{s}+O(s^{-2}).
\end{align*}
\end{lemma}
\noindent
Thus normalized readouts remove the direct radial CE signal, while pre-norm recurrence can still update scale at leading order.
The lemmas identify a local mechanism, not an unconditional theorem of explosion.
The sign and persistence of $a_{\rad}(u_k)$ are learned empirical properties.
If the learned block has persistent positive radial velocity, however, the local CE path through a normalized readout has no direct way to oppose it.
Combining the two statements gives the local scale-control mismatch
\begin{equation}
    \langle \nabla_{H_k}\mathcal{L}_k,H_k\rangle=O(\epsilon/s_k^2),
    \qquad
    s_{k+1}-s_k=a_{\rad}(u_k)+O(s_k^{-1}),
\end{equation}
where $\epsilon$ is the normalizer epsilon; the first term is exact in the epsilon-free idealization and approximate with practical normalizers.
Proof details are in Appendix~\ref{app:proofs}.

\paragraph{Tokenwise normalization and transformer blocks.}
The scalar decomposition keeps notation readable.
In the actual transformer, RMSNorm is applied over channels separately at each token, so the scale-blind direction is one radial degree of freedom per token vector.
The same expansion also applies to a multi-sublayer pre-norm decoder block at leading order: after the first residual addition, later normalizers see $su+O(1)$, so their normalized input differs from $\mathrm{Norm}(u)$ only by $O(s^{-1})$.
Appendix~\ref{app:per-token-norms} reports token-level norm statistics.

\paragraph{Indirect future-loop signals.}
Future losses are not zero-signal.
A later loss can backpropagate through the recurrence and therefore depend indirectly on the earlier scale.
The issue is attenuation: for a fixed future horizon $t$, changing $s_k$ mainly changes the size of later angular steps, $u_{j+1}-u_j=b_\perp(u_j)/s_j+O(s_j^{-2})$.
Under smoothness and bounded-update assumptions on a stable trajectory, this yields
\begin{equation}
    \biggl|\frac{\partial \mathcal{L}_{k+t}}{\partial \log s_k}\biggr|
    \lesssim \frac{C t}{s_k}
\end{equation}
for a constant $C$ depending on those bounds.
We offer this as an explanatory estimate rather than a global-training theorem: it is the fixed-depth reason future CE can be a weak scale-control path at large norm.

\paragraph{Design rule.}
Dense CE trains visible prediction interfaces.
Recurrent scale control requires either making scale visible to at least one loss, for example with raw readouts or an explicit norm penalty, or removing/resetting scale inside the recurrent path, as inter-loop normalization does.
Table~\ref{tab:interventions-main} organizes the interventions used in this paper by which of these roles they play.

\begin{table}[t]
\centering
\small
\caption{\textbf{Scale control and exit training are separate design requirements.} The interventions separate whether CE sees scale, whether scale is controlled by an auxiliary loss, or whether scale is removed from the recurrent path.}
\label{tab:interventions-main}
\begin{tabular}{p{0.30\linewidth}p{0.60\linewidth}}
\toprule
\textbf{Intervention} & \textbf{Role in the framework} \\
\midrule
\multicolumn{2}{l}{\emph{Scale-hidden CE}} \\
Normalized readout & Hides hidden-state scale from the immediate CE loss. \\
Per-loop CE through normalized readouts & Trains intermediate prediction exits, but does not by itself make scale visible. \\
\addlinespace
\multicolumn{2}{l}{\emph{Scale-visible CE}} \\
Raw readout & Makes scale visible to CE by breaking scale invariance. \\
\addlinespace
\multicolumn{2}{l}{\emph{Explicit scale penalty}} \\
Norm penalty & Adds an auxiliary scale-visible objective while keeping the normalized readout. \\
\addlinespace
\multicolumn{2}{l}{\emph{Hybrid readout}} \\
Final-only norm & Exposes intermediate loops through raw readouts but keeps a normalized final interface. \\
\addlinespace
\multicolumn{2}{l}{\emph{Scale-removing recurrence}} \\
Inter-loop normalization & Removes scale from the recurrent input rather than supervising it. \\
\bottomrule
\end{tabular}
\end{table}

Raw readout by itself is not the causal proof.
It is a useful scale-visible intervention, but it also changes logit scale, calibration, output-weight usage, and gradient magnitudes.
The scale-visibility interpretation comes from convergence across controls: raw readouts, explicit norm penalties, final-only normalization, radial-gradient diagnostics, and the recurrent-scale clamp all point to the same active-but-hidden scale direction.

% ============================================================================
\section{Experimental Setup}
\label{sec:setup}

We train autoregressive looped transformers on WikiText-103 \citep{merity2017pointer} with $K=4$ recurrent applications of a shared decoder stack.
Each recurrent step applies the same full stack: 8 pre-norm RMSNorm/SwiGLU transformer layers for the 44M model ($d=512$, 8 heads), or 12 layers for the 129M model ($d=768$, 12 heads).
We denote the stack output after loop $k$ by $H_k$.
The output projection $W_{\mathrm{out}}$ is shared across exits.
A normalized readout computes logits from $\mathrm{RMSNorm}(H_k)$, while a raw readout computes logits directly from $H_k$.
The controlled ablations remove inter-loop normalization, so the unnormalized $H_k$ is fed into the next recurrent application and hidden-state scale remains in the recurrent path.
We train both sizes for 4 epochs over 3 seeds.

The core ablation crosses loss placement and readout:
\begin{equation}
    \{\text{terminal-only CE},\text{ per-loop CE}\}\times
    \{\text{RMSNorm readout},\text{ raw readout}\}.
\end{equation}
Per-loop CE averages the $K$ token-level losses.
We also include two controls: \emph{final-only norm}, which uses raw intermediate readouts but an RMSNorm final readout, and \emph{norm penalty}, which keeps RMSNorm readouts but adds
$\lambda K^{-1}\sum_k \mathbb{E}\|H_k\|_{\rms}^2$ with $\lambda=0.01$.
We report cross-entropy and perplexity (PPL), where PPL is the exponentiated token-level CE.
Reported norms are tokenwise Euclidean $\|H\|_2$ averaged over tokens unless stated otherwise.
We use separate evaluation harnesses for different questions and compare absolute PPL only within a table or figure: Table~\ref{tab:main} uses the full-validation training-time harness, Table~\ref{tab:depth} uses a fixed-depth variable-depth slice harness, and Appendix~\ref{app:halt-protocol} uses a separate timed-slice halting protocol.
Appendix~\ref{app:repro} gives optimizer, tokenizer, harness, and runtime details.

% ============================================================================
\section{Dense CE Does Not Control Hidden-State Scale}
\label{sec:drift}

If norm drift were mainly caused by sparse supervision, adding CE at every loop should control it.
If the problem is scale invisibility, per-loop CE should still fail when every readout is scale-invariant, while scale-visible objectives should control the norm.
Table~\ref{tab:main} supports the second explanation.

\begin{table}[t]
\centering
\small
\caption{\textbf{Dense CE does not control scale when every readout is scale-invariant.} WikiText-103 validation PPL and final-loop hidden-state norm at training depth $K=4$, evaluated with the training-time full-validation harness. Values are mean $\pm$ std over 3 seeds unless a std is omitted. Within-table comparisons are like-for-like; absolute PPL differs from Tables~\ref{tab:depth}, \ref{tab:halting-main}, and~\ref{tab:endpoint-scale-full} because those use shorter slice harnesses (see Section~\ref{sec:setup}).}
\label{tab:main}
\begin{tabular}{llrrrr}
\toprule
 & & \multicolumn{2}{c}{44M} & \multicolumn{2}{c}{129M} \\
\cmidrule(lr){3-4}\cmidrule(lr){5-6}
\textbf{Loss} & \textbf{Readout} & $\|H_K\|_2$ & PPL & $\|H_K\|_2$ & PPL \\
\midrule
Terminal & RMSNorm & $5{,}674{\pm}393$ & $5.40$ & $5{,}997{\pm}25$ & $4.88$ \\
Terminal & Raw & $15{\pm}2$ & $5.30$ & $24{\pm}3$ & $4.92$ \\
\textbf{Per-loop} & \textbf{RMSNorm} & {\boldmath$39{,}207{\pm}19{,}891$} & $6.04$ & {\boldmath$56{,}051{\pm}23{,}993$} & $5.35$ \\
Per-loop & Raw & $44{\pm}1$ & $5.38$ & $77{\pm}4$ & $4.92$ \\
Per-loop & Final-only norm & $57{\pm}2$ & $5.42$ & $83{\pm}10$ & $4.93$ \\
Terminal + penalty & RMSNorm & $21$ & $5.28$ & $27{\pm}0$ & $4.81$ \\
Per-loop + penalty & RMSNorm & $17{\pm}0$ & $5.44$ & $22{\pm}0.3$ & $4.91$ \\
\bottomrule
\end{tabular}
\end{table}

The key negative result is per-loop + RMSNorm: it receives a CE loss at every loop, yet reaches final-loop norms of $39{,}207$ at 44M and $56{,}051$ at 129M.
Raw readouts keep norms in the tens under both terminal-only and per-loop CE.
The norm-penalty controls show that raw readout is not the only solution: even with the same RMSNorm readout, an explicit scale-visible loss collapses norms to 17--27.
Final-only norm gives the practical hybrid: intermediate loops receive scale-sensitive raw losses, while the final interface remains normalized.
Figure~\ref{fig:decomp} summarizes why the two interventions are not substitutes.
Per-loop CE makes exits usable, but it does not by itself control scale.
Scale visibility controls scale, but terminal-only raw readout still fails as an early-exit model.
The desired corner requires both an exit-training signal and a scale-control signal.

\paragraph{The drift is token-level, not just a mean shift.}
Because RMSNorm is tokenwise, a global average could hide the wrong failure mode.
Appendix~\ref{app:per-token-norms} reports per-token final-loop norms on a held-out slice.
The same pattern holds below the mean: terminal-only RMSNorm drift is fairly uniform, per-loop RMSNorm drift is heavy-tailed with top-percentile tokens reaching $10^5{+}$, and raw readouts keep the whole distribution in a narrow range.

% ============================================================================
\section{Mechanism Diagnostics}
\label{sec:diagnostics}

The ablation matches the framework, but we also test the local mechanism directly.
For the immediate readout gradient, RMSNorm readouts have radial-gradient fractions around $10^{-8}$, while raw readouts have radial fractions around $10^{-2}$ (Table~\ref{tab:radial}).
The final-only norm hybrid has a raw-like radial gradient at intermediate loops and a normalized final readout, as designed.
At 1.4B, Appendix~\ref{app:scale14b} repeats the same scale-intervention diagnostic as a sanity check.

\begin{figure}[t]
\centering
\includegraphics[width=\linewidth]{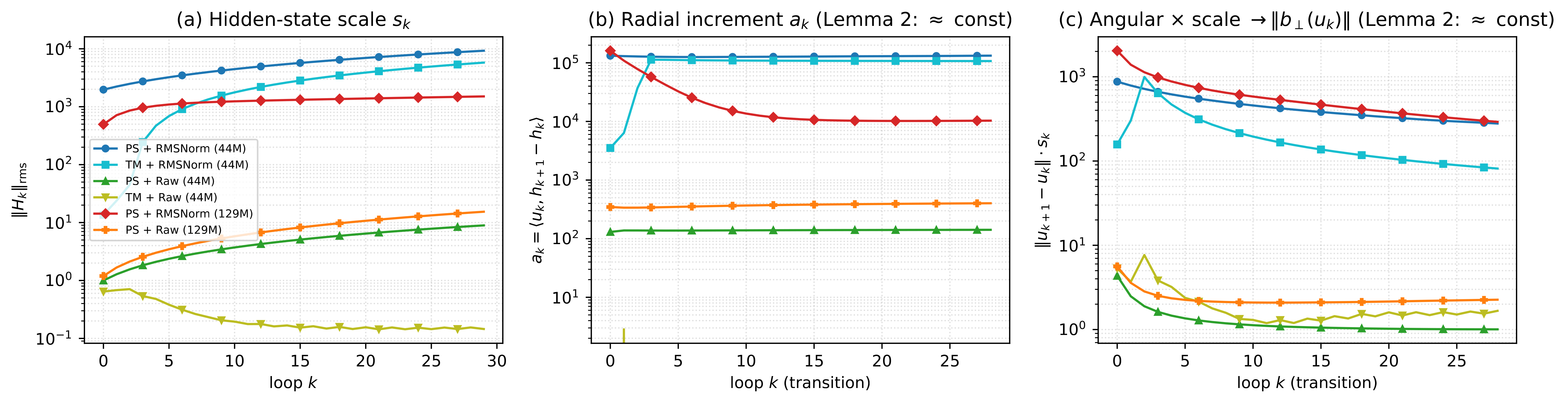}
\caption{\textbf{Trained checkpoints enter the slow-angular-motion regime predicted by the scale expansion.}
(a) Hidden-state RMS scale over 30 loops.
(b) Euclidean radial residual inner product $\tilde a_k=\langle u_k,F(H_k)-H_k\rangle$, whose scaled value $\tilde a_k/d$ predicts the RMS-scale increment.
(c) $\|u_{k+1}-u_k\|_{\rms}s_k$, whose leading-order value is $\|b_\perp(u_k)\|_{\rms}$.}
\label{fig:lemma2-main}
\end{figure}

\paragraph{Active radial update.}
Lemma~\ref{lemma:expansion} predicts that, once the model is in the large-scale regime, the RMS-scale increment is approximately
$a_{\rad,k}=\langle u_k,F(H_k)-H_k\rangle/d$, while angular motion shrinks like $1/s_k$.
Figure~\ref{fig:lemma2-main} checks this directly by running trained checkpoints for 30 loops on held-out data.
The important empirical fact is not the Taylor expansion itself; it is that the learned residual updates enter a stable regime where a persistent radial component accounts for the observed scale motion.

The measured residual update magnitude also separates sharply by readout.
Across the diagnostic checkpoints, $\|b\|_{\rms}$ lies in $[0.07,0.51]$ for raw readouts but in $[20.4,258.3]$ for normalized readouts.
For example, the 44M per-loop + RMSNorm checkpoint has $\tilde a\approx132{,}000$, predicting an RMS-scale increment of about $258$ per loop, within $3\%$ of the measured value once the transient settles.
The corresponding per-loop + raw checkpoint predicts an increment of only $0.27$.
The full decomposition is in Appendix~\ref{app:residual-update}; the normalized-readout checkpoints learn much larger radial residual updates than the raw-readout checkpoints.

\begin{table}[t]
\centering
\small
\caption{\textbf{Readout normalization removes the local radial gradient.}
$r_k$ is the fraction of the readout gradient aligned with hidden-state scale at loop $k$; $g_K=\partial\mathcal{L}/\partial\log\|H_K\|$ is the signed final-loop radial derivative.
Raw intermediate readouts restore a scale-sensitive signal; final-only norm restores it only before the final normalized interface.}
\label{tab:radial}
\begin{tabular}{llrrrr}
\toprule
\textbf{Model} & \textbf{Readout} & $\|H_K\|_2$ & $r_1$ & $r_K$ & $g_K$ \\
\midrule
44M & RMSNorm & $62{,}847$ & $5.0{\times}10^{-9}$ & $4.7{\times}10^{-9}$ & $4.0{\times}10^{-10}$ \\
44M & Raw & $41$ & $6.3{\times}10^{-2}$ & $7.2{\times}10^{-2}$ & $8.2{\times}10^{-2}$ \\
44M & Final-only norm & $54$ & $6.4{\times}10^{-2}$ & $1.2{\times}10^{-8}$ & $-3.9{\times}10^{-8}$ \\
129M & RMSNorm & $26{,}045$ & $8.8{\times}10^{-9}$ & $7.5{\times}10^{-9}$ & $-3.1{\times}10^{-9}$ \\
129M & Raw & $70$ & $6.0{\times}10^{-2}$ & $4.7{\times}10^{-2}$ & $1.4{\times}10^{-1}$ \\
129M & Final-only norm & $91$ & $5.9{\times}10^{-2}$ & $9.5{\times}10^{-9}$ & $4.9{\times}10^{-8}$ \\
\bottomrule
\end{tabular}
\end{table}

We also run a causal clamp.
After loop 1, we rescale each token's hidden state back to its loop-1 RMS scale before both decoding and recurrence.
This removes recurrent scale growth while preserving direction changes.
Table~\ref{tab:radial-clamp} shows that this changes CE very little.

\begin{table}[t]
\centering
\small
\caption{\textbf{Clamping recurrent scale has little CE cost.} Norm ratio is clamped $K=4$ norm divided by unclamped $K=4$ norm; $\Delta$CE is clamped minus unclamped CE. Means over 3 seeds.}
\label{tab:radial-clamp}
\begin{tabular}{llrr}
\toprule
\textbf{Readout} & \textbf{Model} & \textbf{$K=4$ norm ratio} & \textbf{$\Delta$CE} \\
\midrule
RMSNorm & 44M & 0.40 & $+0.0004$ \\
RMSNorm & 129M & 0.55 & $+0.0005$ \\
Raw & 44M & 0.53 & $+0.0055$ \\
Raw & 129M & 0.46 & $+0.0015$ \\
Final-only norm & 44M & 0.41 & $+0.0035$ \\
Final-only norm & 129M & 0.42 & $+0.0027$ \\
\bottomrule
\end{tabular}
\end{table}

The clamp does not prove every radial residual component is useless.
It shows that accumulated scale growth itself is predictively cheap to remove under this evaluation, consistent with the visibility--activity mechanism.
Appendices~\ref{app:per-token-norms}--\ref{app:residual-update} give token-level norm statistics and exact decomposition values.

% ============================================================================
\section{Scale Control Shifts the Variable-Depth Curve}
\label{sec:depth}

We next ask whether scale control changes how models use recurrent depth at inference time.
The result is not a generic speedup claim: a model can look fast simply because it is already $K$-invariant.
The stronger comparison is the fixed-depth PPL--compute frontier.

\begin{table}[t]
\centering
\small
\caption{\textbf{Scale-controlled per-loop models improve with depth and stay lower-PPL.}
Variable-depth inference on per-loop-loss models, mean PPL over 3 seeds. The $K=1$ and $K=4$ columns use the fixed-depth variable-depth slice harness; the dynamic-loop column summarizes a separately calibrated timed-slice halting protocol detailed in Appendix~\ref{app:halt-protocol}. Terminal-only variable-depth baselines (129M) are in Appendix~\ref{app:variable-depth}.}
\label{tab:depth}
\begin{tabular}{llrrrr}
\toprule
\textbf{Readout} & \textbf{Model} & $K=1$ & $K=4$ & $\Delta$PPL & Dynamic avg.\ loops \\
\midrule
RMSNorm & 44M & 5.58 & 5.59 & $+0.01$ & 1.00 \\
RMSNorm & 129M & 4.98 & 4.97 & $-0.01$ & 1.00 \\
Raw & 44M & 5.14 & 4.94 & $-0.20$ & 2.16 \\
Raw & 129M & 4.68 & 4.55 & $-0.13$ & 1.76 \\
Final-only norm & 44M & 5.20 & 5.00 & $-0.20$ & 1.78 \\
Final-only norm & 129M & 4.70 & 4.55 & $-0.15$ & 1.88 \\
Norm penalty & 44M & 5.22 & 5.00 & $-0.22$ & 2.60 \\
Norm penalty & 129M & 4.67 & 4.53 & $-0.14$ & 2.56 \\
\bottomrule
\end{tabular}
\end{table}

RMSNorm per-loop models are nearly $K$-invariant: at both sizes they can halt at $K=1$ under a 1\% PPL budget.
That produces high apparent speedup, but the model has little measurable use for additional loops.
This is the behavior Lemma~\ref{lemma:expansion} predicts at large scale: with final-loop norms in the tens of thousands, angular updates shrink like $1/s_k$, so additional loops barely change the normalized state the readout sees.
Norm drift therefore does not show up as a worse $K=4$ loss; its cost is the depth dimension itself.
Raw, final-only norm, and norm penalty all recover depth use: they remain usable at $K=1$ but improve with more loops.

\paragraph{High halting speedup can reflect $K$-invariance.}
We also benchmark a sequence-level logit-margin halting rule, calibrated to keep dynamic perplexity within $1\%$ of always running $K=4$.
The full dynamic-halting table is in Appendix~\ref{app:halt-protocol}; the result shows why a single speedup ratio is not the right endpoint for this paper.
The RMSNorm model exits at $K=1$ for every sequence and therefore appears fastest, but Table~\ref{tab:depth} shows that this happens because the model is already nearly $K$-invariant.
The scale-controlled rows spend more loops because additional loops actually improve perplexity.

\begin{figure}[t]
\centering
\includegraphics[width=0.78\linewidth]{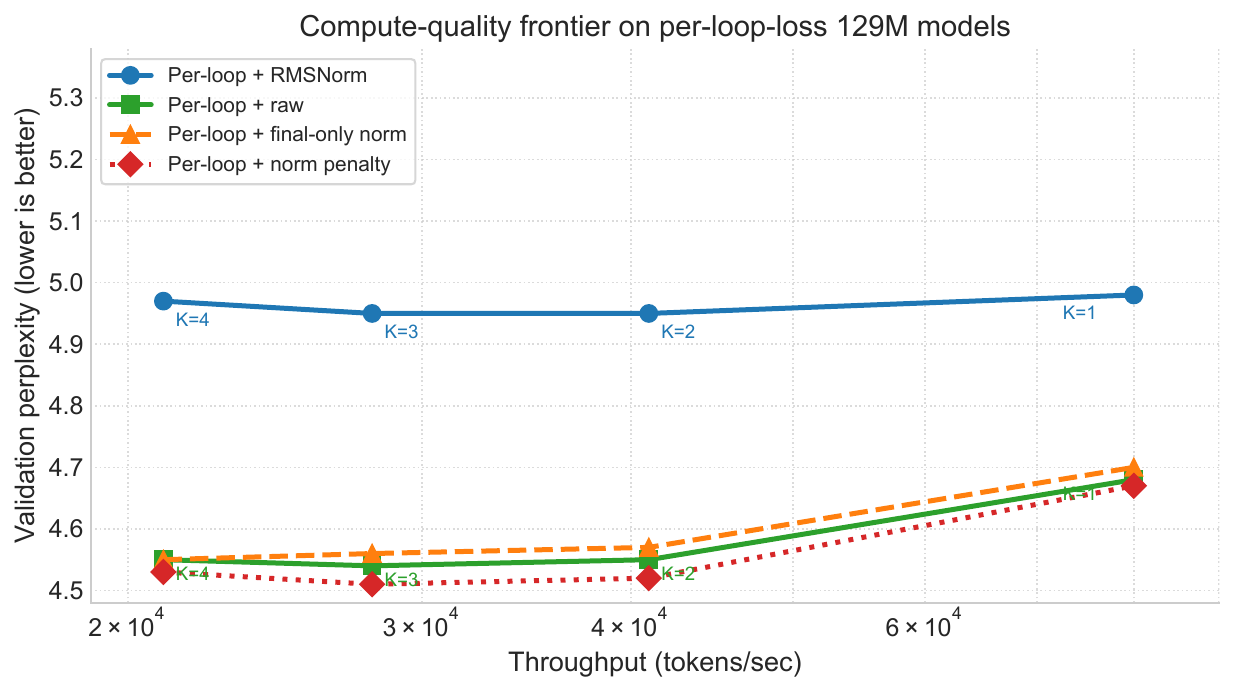}
\caption{\textbf{Compute--quality frontier on per-loop-loss 129M models.}
Each curve sweeps $K\in\{1,2,3,4\}$.
Raw, final-only norm, and norm penalty occupy the lower-PPL region; RMSNorm is above them at every measured operating point.}
\label{fig:pareto}
\end{figure}

The Pareto-relevant comparison is therefore the joint PPL--throughput frontier (Figure~\ref{fig:pareto}), not the speedup ratio in isolation.
For example, raw readout at $K=1$ has the same measured throughput as RMSNorm at $K=1$ but substantially lower PPL.

At equal throughput ($K=1$, about 80k tokens/s), raw readout is 0.30 PPL lower than RMSNorm.
At full depth ($K=4$, about 21k tokens/s), the gap is 0.42 PPL.
The conclusion is a compute--quality claim, not a higher-speedup claim: in this benchmark, scale-controlled per-loop variants preserve usable exits while shifting the measured frontier below the $K$-invariant RMSNorm baseline.

\paragraph{A simple practical intervention: scale-visibility penalty.}
For practitioners who want to keep normalized readouts, the norm penalty is the most direct intervention in our study.
The implementation is a one-line auxiliary loss,
$\lambda K^{-1}\sum_k \mathbb{E}\|H_k\|_{\rms}^2$, using hidden states already computed during the forward pass; its overhead is a reduction over activations, negligible compared with another recurrent block.
Table~\ref{tab:interventions-main} identifies it as an explicit scale-visible objective, Table~\ref{tab:main} shows that it collapses final-loop norms to the same range as raw readouts, and Table~\ref{tab:depth} shows that it preserves usable exits while recovering depth-dependent PPL improvement.
We do not claim this penalty is optimal.
Its role is simpler: it is a drop-in baseline that separates scale control from raw-readout confounds and is worth including whenever normalized readouts are kept but recurrent scale remains active.

Final-only normalization gives the complementary interface result: it uses raw intermediate readouts to control recurrence, but returns to a normalized final readout for the endpoint used by a fixed-depth model.
Together these controls make the conclusion less about one preferred readout and more about the underlying requirement.
Adaptive-depth looped models need trained exits and a way to control active recurrent scale.
Raw readout, an explicit penalty, and scale-removing recurrence are different engineering points on that same requirement; Appendix~\ref{app:mitigations} reports additional mitigation-style controls.

% ============================================================================
\section{Related Work}
\label{sec:related}

\paragraph{Recurrent-depth and looped transformers.}
Universal Transformers \citep{dehghani2019universal} combine depth-wise weight sharing with adaptive computation.
Reasoning with Latent Thoughts \citep{saunshi2025latentthoughts} and Loop, Think, \& Generalize \citep{kohli2026loopthink} show that looped transformers can trade repeated latent computation for feedforward depth on reasoning-style tasks.
Huginn \citep{geiping2025huginn} scales recurrent-depth language modeling and frames latent reasoning as a third test-time compute axis.
Ouro \citep{zhu2025ouro} pretrains a looped language model with inter-loop normalization and entropy-regularized adaptive exit.
LoopFormer \citep{jeddi2026loopformer} trains budget-conditioned looped models with shortcut modulation.
Elastic Looped Transformer (ELT) \citep{goyal2026elt} studies visual generation with looped transformers and uses intra-loop self-distillation to make intermediate loops usable.
Parcae \citep{prairie2026parcae} studies stable looped language-model scaling through a dynamical-systems view and spectral constraints on looped updates.
Recent mechanistic analyses \citep{blayney2026mechanistic,labovich2026stability} study cyclic fixed points, stages of inference, normalization, and stability in looped models.
Our focus is complementary: we study how the supervised output readout determines whether CE can see recurrent hidden-state scale.

\paragraph{Adaptive computation and early exit.}
Adaptive Computation Time \citep{graves2016adaptive}, PonderNet \citep{banino2021pondernet}, and depth-adaptive Transformers \citep{elbayad2020depth} all motivate spending different amounts of computation on different examples.
Our setting differs because every exit is another application of the same recurrent block, so the hidden state used for an early prediction is also the state that would be fed into later computation.
This makes readout geometry part of the adaptive-compute problem.

\paragraph{Normalization and recurrent scale.}
RMSNorm \citep{zhang2019rmsnorm} explicitly provides re-scaling invariance.
LayerNorm placement \citep{xiong2020layernorm} changes transformer optimization, while NormFormer \citep{shleifer2021normformer} and DeepNet \citep{wang2022deepnet} add normalization or residual scaling to improve fixed-depth pretraining.
TaperNorm \citep{kanavalau2026tapernorm} is the closest output-side comparison: it studies removing normalization while maintaining output scale anchoring and avoiding logit chasing in standard transformers.
Our distinction is the recurrent interface.
Output normalization can stabilize fixed-depth transformers; in a looped model, however, the same hidden state is reused by future loops.
Thus the readout is not only an output calibration choice: it determines whether CE can supervise an active recurrent state variable.
When every CE signal for that recurrent state passes through a scale-invariant interface, scale can become an underconstrained recurrent degree of freedom.
Classical recurrent stability work \citep{pascanu2013difficulty} studies exploding and vanishing dynamics; our results show a compatible lever: make the loss itself sensitive to recurrent scale, or remove scale from the recurrent state.

\paragraph{Concurrent work.}
Several concurrent 2026 works study the stability and supervision of looped models from directions complementary to ours.
On the dynamics side, the Fully Looped Transformer \citep{fu2026fullylooped} attributes training instability at high loop counts to gradient oscillation and residual explosion and responds architecturally, while STARS \citep{yang2026stars} regularizes the Jacobian spectral radius so that latent states approach stable fixed points at large test-time depth.
Both intervene on the recurrent dynamics directly; we instead ask which recurrent variables the supervised readout lets any loss control, and show that scale can drift under dense supervision precisely because every supervised interface hides it.
On the supervision side, RLTT \citep{williams2026rltt} and LoopRPT \citep{tang2026looprpt} find that reinforcement objectives on looped models improve when credit is assigned to intermediate latent states rather than only the final state.
This matches our observation that loss placement determines which parts of the latent trajectory are trained; the visibility question we raise, namely what each per-step signal can see through the readout, applies to those objectives as well.
Finally, concurrent scaling studies \citep{schwethelm2026isodepth,lee2026sparse} quantify when looping is worth its training compute and identify loop boundaries as natural early-exit points, reinforcing the usable-intermediate-exit requirement that motivates our variable-depth evaluation.

% ============================================================================
\section{Limitations}
\label{sec:limitations}

Our conclusions are deliberately scoped to the settings studied in this paper. The main controlled evidence comes from 44M- and 129M-parameter looped language models trained on WikiText-103, where inter-loop normalization is removed so that recurrent scale remains active in the recurrent state. Architectures with inter-loop normalization, gating, clipping, or other recurrent-state controls may avoid this failure mode by construction. Appendix~\ref{app:scale14b} reports a separate 1.4B Ouro-scale experiment trained on FineWeb as a scale sanity check, but not as a full controlled ablation.

The proposed mechanism is also local rather than fully global. Our analysis characterizes the immediate cross-entropy signal seen through the readout and explains why recurrent scale can become weakly supervised under scale-invariant readouts. It does not rule out indirect future-loop gradients or fully characterize optimization dynamics at larger scales, longer training horizons, or different data regimes.

Finally, our adaptive-depth evaluations use teacher-forced language-model scoring rather than production autoregressive generation. The resulting perplexity--compute curves provide controlled comparisons across recurrent-depth strategies, but serving behavior may differ under key-value (KV) caching, large-batch inference, or deployment workloads. Exploring these settings, extending the analysis to larger-scale models, and studying additional recurrent-state variables beyond scale are important directions for future work.

% ============================================================================
\section{Conclusion}
\label{sec:conclusion}
Looped language models make each hidden state serve two roles: it is both a prediction interface and the input to future recurrent computation.
Dense cross-entropy trains the prediction interface, but it does not automatically control every recurrent variable carried by that state.
Scale makes this failure concrete.
With a scale-invariant readout, the immediate cross-entropy loss is largely insensitive to hidden-state scale, while pre-norm residual recurrence continues to carry and update it.
Our ablations separate these two requirements: per-loop cross-entropy makes intermediate exits usable, but recurrent scale control requires either making hidden-state scale visible to a loss or removing it from the recurrent path.

Raw readouts, explicit norm penalties, hybrid readouts, and scale-removing recurrence are different ways to satisfy the scale-control requirement.
For practitioners who want to keep normalized readouts, the explicit norm penalty is a drop-in scale-control baseline with negligible overhead relative to an extra recurrent step.
The main conclusion is not that normalized readouts are inherently problematic.
Rather, active recurrent variables require either visibility to training or explicit removal from the recurrent dynamics.
\FloatBarrier

% ============================================================================
\section*{Acknowledgments}

We thank Chris Thomas for insightful conversations and feedback on the manuscript.
The authors acknowledge Advanced Research Computing at Virginia Tech (\url{https://arc.vt.edu/}) for providing computational resources and technical support that have contributed to the results reported within this paper.

\bibliographystyle{plainnat}
\bibliography{references}

\appendix

% appendix begins (\appendix is issued in main.tex)

% ============================================================================
\section{Proof Details for Section~\ref{sec:mechanism}}
\label{app:proofs}

\paragraph{Visibility lemma.}
Define $\phi(t)=\mathcal{L}(e^tH)$.
If the readout is scale-invariant, $r(e^tH)=r(H)$ for all $t$, so $\phi(t)$ is constant.
Therefore $\phi'(0)=0$.
By the chain rule,
\begin{equation}
    \phi'(0)=\left\langle \nabla_H \mathcal{L}(H), H\right\rangle,
\end{equation}
which proves Lemma~\ref{lemma:radial}.
With RMSNorm/LayerNorm epsilon terms, scale invariance is approximate rather than exact; at large scale the epsilon contribution is suppressed, giving the $O(\epsilon/s^2)$ radial term used in the main text.

\paragraph{Activity lemma.}
For $H=su$ with $\|u\|_{\rms}=1$ and $\|v\|_{\rms}^2=d^{-1}\|v\|_2^2$, ignoring normalizer epsilons gives
\begin{equation}
    F(su)=su+b(u).
\end{equation}
Decompose the residual update into radial and orthogonal parts,
\begin{equation}
    b(u)=a_{\rad}(u)u+b_\perp(u),
    \qquad
    a_{\rad}(u)=\frac{\langle u,b(u)\rangle}{d},
    \qquad
    \langle u,b_\perp(u)\rangle=0 .
\end{equation}
Then
\begin{equation}
    \|su+b(u)\|_{\rms}^2
    =s^2+2s\,a_{\rad}(u)+\|b(u)\|_{\rms}^2 .
\end{equation}
Taking the square root and using $\|b(u)\|_{\rms}=O(1)$ in $s$ gives
\begin{equation}
    \|F(su)\|_{\rms}=s+a_{\rad}(u)+O(s^{-1}).
\end{equation}
Substituting this denominator into
\begin{equation}
    \frac{F(su)}{\|F(su)\|_{\rms}}
    =
    \frac{(s+a_{\rad}(u))u+b_\perp(u)}
         {s+a_{\rad}(u)+O(s^{-1})}
\end{equation}
gives
\begin{equation}
    \frac{F(su)}{\|F(su)\|_{\rms}}
    =
    u+\frac{b_\perp(u)}{s}+O(s^{-2}),
\end{equation}
which proves Lemma~\ref{lemma:expansion}.

% ============================================================================
\section{1.4B Scale Sanity Check}
\label{app:scale14b}

The controlled 44M and 129M experiments isolate the mechanism.
We also ran a near-matched-token Ouro-scale check to ask whether the readout-side diagnostic survives in a larger implementation trained on a modern corpus.
This appendix reports that check.
It should be read as scale evidence only: it is not a full 2$\times$2 ablation, not a seeded causal comparison, and not an adaptive-depth speedup result.

We train two Ouro-scale 1.4B models on FineWeb \citep{penedo2024fineweb}: a baseline with the standard normalized readout and a no-readout-normalization variant.
Both are evaluated after 50{,}000 steps, corresponding to near-matched training token counts: 1.321B tokens for the baseline and 1.303B tokens for the no-readout-norm checkpoint.

\begin{figure}[t]
\centering
\includegraphics[width=0.93\linewidth]{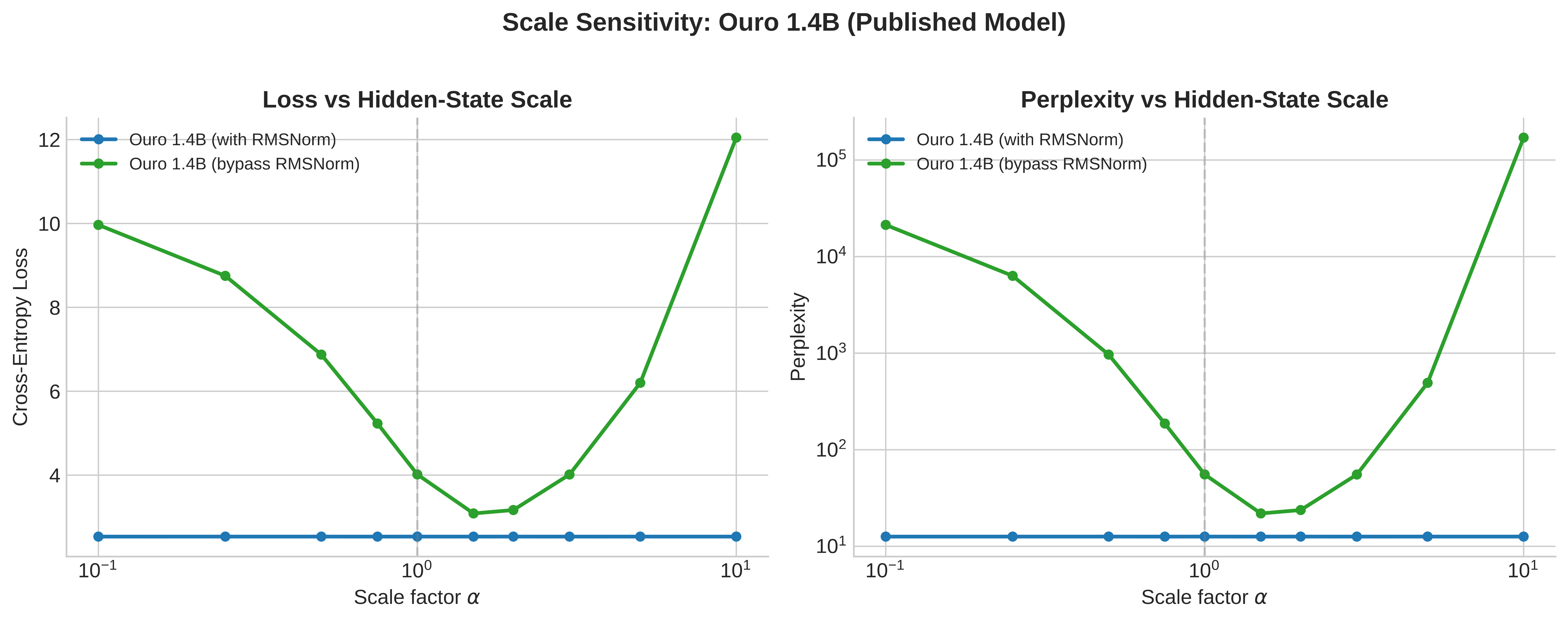}
\caption{\textbf{A normalized readout hides hidden-state scale at billion-parameter scale.} We scale the final hidden state of the published Ouro 1.4B checkpoint by $\alpha \in [0.1, 10]$ before the readout and measure cross-entropy. With RMSNorm, CE is essentially unchanged across a $100\times$ range, consistent with Lemma~\ref{lemma:radial}. Without RMSNorm, the same hidden states produce a sharp scale-sensitive curve. This is a direct measurement on a trained checkpoint, not a retraining experiment.}
\label{fig:scale}
\end{figure}

\subsection{Mechanism Diagnostic at Near-Matched Tokens}
\label{sec:scale14b-mechanism}

\begin{figure}[t]
\centering
\includegraphics[width=0.7\linewidth]{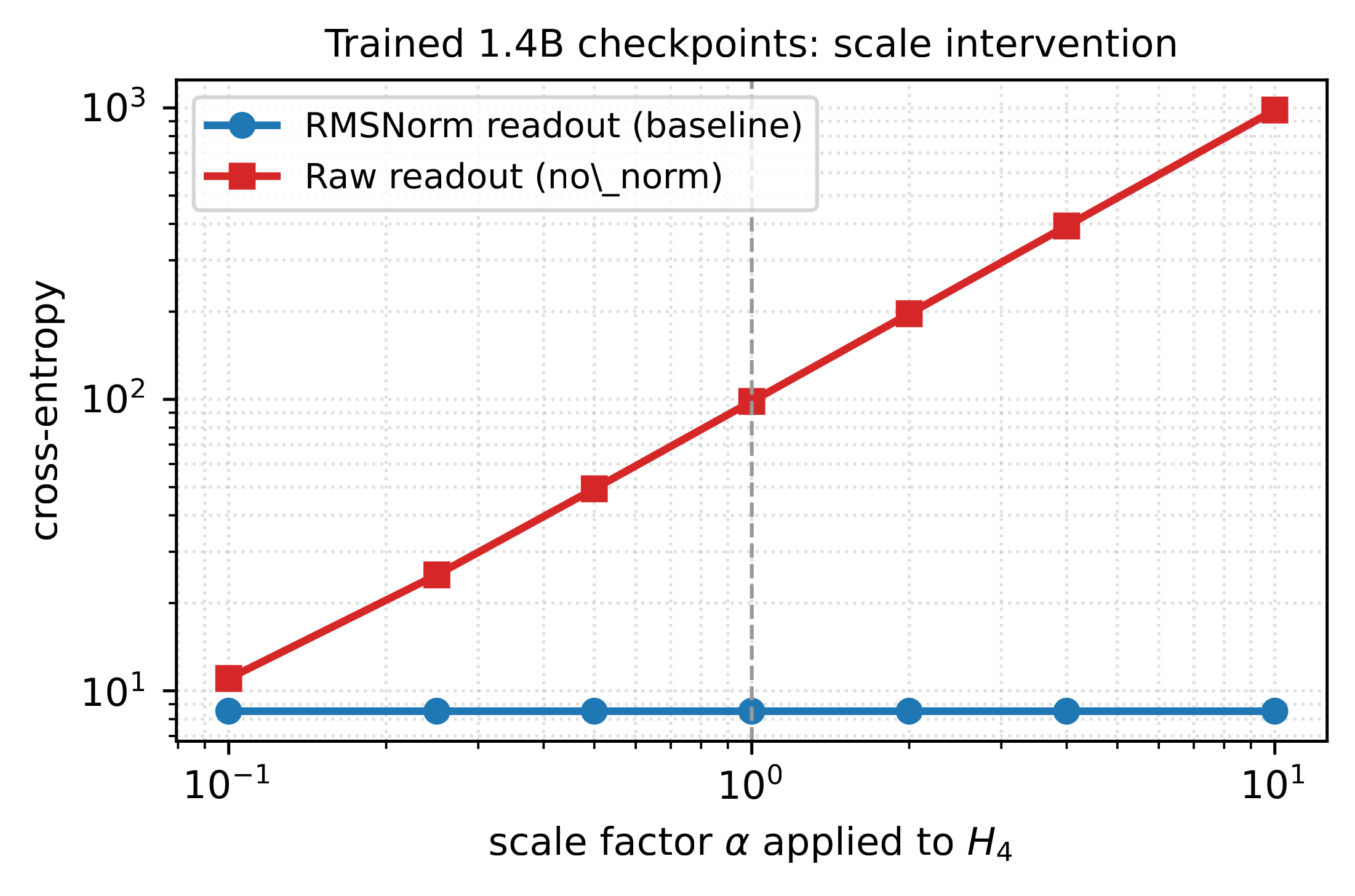}
\caption{\textbf{Direct scale intervention on trained 1.4B checkpoints.} We multiply the final hidden state by $\alpha$ before the readout and measure cross-entropy. The baseline (RMSNorm readout) is flat across the whole $\alpha$ range. The no-readout-norm variant produces a sharp scale-sensitive curve. This is the same intervention as Figure~\ref{fig:scale}, but on our own trained checkpoints rather than the published Ouro: the readout blind spot persists at billion-parameter scale, and removing the readout normalization restores the loss's view of hidden-state scale.}
\label{fig:scale_trained}
\end{figure}

We first run the readout-scale diagnostic on the two trained 1.4B checkpoints themselves.
Ouro's remote-code training path computes the loss from an exit-weighted mixture of per-loop logits when labels are supplied, so we separate the training-objective CE from final-readout diagnostics.
Table~\ref{tab:scale14b-mech} shows that the near-matched-token checkpoints have nearly identical training-objective CE and logit statistics on a WikiText-103 subset, while the final readout's radial gradient differs by roughly three orders of magnitude.
The normalized baseline remains essentially flat under final-hidden rescaling; the no-readout-norm checkpoint is strongly scale-sensitive (Figure~\ref{fig:scale_trained}).

\begin{table}[h]
\centering
\small
\caption{Near-matched-token 1.4B mechanism diagnostic on 4{,}096 WikiText-103 validation tokens. Training CE and logit statistics are computed on Ouro's label path, which uses exit-weighted logits. Final $r_K$ is the final-loop radial-gradient fraction. Radial gradient and the scale sweep are diagnostic final-readout measurements and are not meant to equal the training-objective CE.}
\label{tab:scale14b-mech}
\begin{tabular}{lrrrrr}
\toprule
\textbf{Checkpoint} & \textbf{Tokens} & \textbf{Train CE} & \textbf{Entropy} & \textbf{Final $r_K$} & \textbf{Diagnostic final CE} $\alpha=0.1{\to}1$ \\
\midrule
Baseline & 1.321B & 4.987 & 5.30 & $3.96{\times}10^{-5}$ & $8.50{\to}8.50$ \\
No readout norm & 1.303B & 4.999 & 5.27 & $6.80{\times}10^{-2}$ & $11.02{\to}98.36$ \\
\bottomrule
\end{tabular}
\end{table}

At 1.4B parameters, the output interface behaves as predicted: normalized readouts hide the final hidden-state scale from CE, while removing the final readout norm restores a large radial signal.
The scale-sweep column intentionally reports \emph{final-readout} CE rather than training-objective CE; for Ouro this differs from the label-path objective because the latter uses exit-weighted logits.
The similar training-objective entropy and top-logit margin (0.91 vs.\ 0.90, not shown) make a simple logit-chasing explanation less likely, but the comparison remains a scale diagnostic rather than a main ablation.

\subsection{Downstream Behavior at Near-Matched Tokens}
\label{sec:scale14b}

We next evaluate full validation splits from nine multiple-choice tasks using continuation likelihood: BoolQ, WinoGrande, CommonsenseQA, SciQ, PIQA, ARC-Easy, ARC-Challenge, HellaSwag, and OpenBookQA.
The resulting comparison covers 19{,}169 paired examples.

\begin{table}[h]
\centering
\small
\caption{Near-matched-token downstream evaluation at 1.4B parameters. Both models are evaluated after 50{,}000 training steps: baseline at 1.321B tokens, no-readout-norm at 1.303B tokens. Accuracy metrics are macro averages over nine full validation splits. Score metrics are micro-averages over 19{,}169 paired examples using length-normalized continuation scores; 95\% confidence intervals (CIs) are paired bootstrap intervals.}
\label{tab:scale14b}
\begin{tabular}{lrrrr}
\toprule
\textbf{Metric} & \textbf{Baseline} & \textbf{No norm} & \textbf{Change} & \textbf{95\% CI} \\
\midrule
Raw accuracy macro (\%) & 32.55 & 32.39 & $-0.16$ & -- \\
Length-normalized accuracy macro (\%) & 34.72 & 34.56 & $-0.16$ & -- \\
Candidate negative log-likelihood (NLL) $\downarrow$ & 1.475 & \textbf{1.419} & {\boldmath$-0.056$} & $[-0.066,-0.047]$ \\
Gold-vs-distractor margin $\uparrow$ & $-0.484$ & {\boldmath$-0.443$} & {\boldmath$+0.040$} & $[+0.025,+0.055]$ \\
Gold option probability (\%) $\uparrow$ & 31.66 & \textbf{32.14} & {\boldmath$+0.47$} & $[+0.27,+0.69]$ \\
\bottomrule
\end{tabular}
\end{table}

Argmax accuracy is effectively tied, while the paired score metrics favor no-readout-norm: it assigns lower length-normalized NLL to candidate continuations, increases the gold-vs-distractor margin, and assigns higher probability to the gold answer.
The interpretation is limited: accuracy is similar, while likelihood-style metrics modestly favor the no-readout-norm checkpoint at this early training budget.

\subsection{Fixed-K Inference: Iso-\texorpdfstring{$K$}{K} Quality}
\label{sec:fixedk-14b}

We measure wall-clock throughput and per-token cross-entropy at fixed $K \in \{1,2,3,4\}$ on the same step-50{,}000 checkpoints, using a held-out FineWeb-tail slice.
At matched $K$, throughput is essentially identical across readouts; the normalized readout is not an appreciable per-loop runtime cost.

\begin{table}[h]
\centering
\small
\caption{Wall-clock throughput at fixed inference depth $K$ for the 1.4B step-50{,}000 checkpoints (single-GPU timing, 128 timed batches, batch size 2, sequence length 2{,}048, FineWeb-tail validation, bfloat16 (bf16)). At equal $K$ the no-readout-norm checkpoint has lower CE than the baseline. Both checkpoints are nearly $K$-invariant on this slice: CE differs by less than $0.01$ nats across $K{=}2{,}3{,}4$, so this is an iso-$K$ scale check rather than an adaptive-depth speedup result.}
\label{tab:fixedk-14b}
\begin{tabular}{llrrr}
\toprule
\textbf{Readout} & \textbf{$K$} & \textbf{CE} & \textbf{PPL} & \textbf{Tokens/s} \\
\midrule
\multirow{4}{*}{Baseline (RMSNorm)}
 & 4 & 6.7839 & 883.5 & 11{,}746 \\
 & 3 & 6.7839 & 883.5 & 15{,}816 \\
 & 2 & 6.7839 & 883.5 & 24{,}082 \\
 & 1 & 6.7892 & 888.2 & 48{,}154 \\
\midrule
\multirow{4}{*}{No readout norm (raw)}
 & 4 & \textbf{6.7678} & \textbf{869.4} & 11{,}817 \\
 & 3 & \textbf{6.7678} & \textbf{869.4} & 15{,}770 \\
 & 2 & \textbf{6.7678} & \textbf{869.4} & 23{,}741 \\
 & 1 & \textbf{6.7719} & \textbf{873.0} & 47{,}468 \\
\bottomrule
\end{tabular}
\end{table}

Both checkpoints are essentially $K$-invariant on this slice (CE varies by less than $0.01$ nats across $K{=}2{,}3{,}4$), which is consistent with Ouro's inter-loop RMSNorm bounding in-loop scale (see also the published-checkpoint analysis in Appendix~\ref{app:ouro}).
Within this regime, the no-readout-norm checkpoint has modestly lower CE/PPL than the baseline at every measured $K$ while running at indistinguishable throughput.
This supports the readout-side scale diagnostic at larger size, but it does not establish a user-facing speedup.

% ============================================================================
\section{Reproducibility Details}
\label{app:repro}

\begin{table}[!htbp]
\centering
\small
\caption{Implementation and evaluation details for the main experiments. Batch and runtime settings vary slightly across cluster continuations for the 1.4B runs; the reported token counts are read from the saved checkpoints used for evaluation.}
\label{tab:repro}
\begin{tabular}{p{0.26\linewidth}p{0.66\linewidth}}
\toprule
\textbf{Component} & \textbf{Setting} \\
\midrule
44M / 129M data & WikiText-103 raw train/validation; regex tokenizer fit from text with maximum vocabulary 20k and minimum token frequency 2; sequence length 256 and stride 128. \\
44M / 129M training & $K=4$ loops; AdamW with learning rate $3{\times}10^{-4}$, weight decay 0.01, gradient clipping 1.0, dropout 0.1; 4 epochs; 3 seeds; bfloat16 (bf16) on CUDA for full runs. Batch size is 64 for 44M and 32 for 129M. \\
44M / 129M evaluation & Table~\ref{tab:main} reports validation at training depth $K=4$ and averages over seeds. Table~\ref{tab:depth} uses the variable-depth evaluation harness on selected 44M and 129M checkpoints; comparisons within each row use the same checkpoint and evaluation code. \\
Radial clamp & Table~\ref{tab:radial-clamp} uses the same per-loop-loss checkpoints as Table~\ref{tab:depth}. We run 20 batches of 8 sequences, 40{,}960 tokens per checkpoint, run loop 1 normally, and then clamp later tokenwise RMS scales to their loop-1 values before decoding and recurrence. \\
Dynamic halting & Sequence-level halting uses logit margin as the confidence score. Thresholds are calibrated on validation to stay within a 1\% PPL budget relative to fixed $K=4$; throughput includes dynamic batching overhead and is measured with a single-GPU timing setup. \\
1.4B training & ByteDance/Ouro-1.4B architecture initialized from scratch and trained on FineWeb sample-10BT with sequence length 2{,}048, bf16, AdamW, learning rate $3{\times}10^{-4}$, minimum learning rate $3{\times}10^{-5}$, weight decay 0.01, gradient clipping 1.0, and 1000 warmup steps. Checkpoints are saved periodically for preemptible jobs. \\
1.4B checkpoints & The evaluated step-50{,}000 checkpoints contain 1.321B observed training tokens for the baseline and 1.303B for the no-readout-norm variant, a 1.4\% difference. \\
1.4B downstream & Full validation splits for BoolQ, WinoGrande, CommonsenseQA, SciQ, PIQA, ARC-Easy, ARC-Challenge, HellaSwag, and OpenBookQA; max length 2{,}048; paired bootstrap intervals over 19{,}169 examples. \\
1.4B timing & Single-GPU timing; bf16; batch size 2; sequence length 2{,}048; 8 warmup batches and 128 timed batches on a held-out FineWeb-tail slice. \\
\bottomrule
\end{tabular}
\end{table}

\FloatBarrier

% ============================================================================
\section{Token-Level Norm Statistics}
\label{app:per-token-norms}

The mean $\|H_K\|$ in Table~\ref{tab:main} aggregates over tokens and can mask per-token outliers, which is a real concern given that real LayerNorm and RMSNorm operate per token.
Table~\ref{tab:per-token} reports the per-token $\|H_K\|_2$ distribution on $20{,}480$ validation tokens for each 2$\times$2 condition (single seed; the across-seed pattern is the same).

\begin{table}[h]
\centering
\small
\caption{\textbf{Per-loop RMSNorm drift is a token-level heavy-tail failure, not just a mean shift.} Per-token final-loop norm distribution (Euclidean $\|H_K\|_2$) on 20,480 WikiText-103 validation tokens. RMSNorm-readout drift is concentrated in a heavy tail of outlier tokens, especially under per-loop CE: the median is $12$--$21\times$ smaller than the mean, with $1\%$ of tokens reaching $10^5{+}$. Raw readouts produce tighter bounded distributions in every condition.}
\label{tab:per-token}
\begin{tabular}{llrrrrr}
\toprule
\textbf{Model} & \textbf{Condition} & \textbf{Mean} & \textbf{Median} & \textbf{p99} & \textbf{Max} & \textbf{Std} \\
\midrule
44M  & Terminal + RMSNorm & 5{,}644 & 5{,}493 & 8{,}768 & 11{,}798 & 899 \\
44M  & Terminal + Raw     & \textbf{12} & \textbf{6} & \textbf{44} & \textbf{259} & \textbf{12} \\
44M  & Per-loop + RMSNorm & 63{,}347 & 2{,}986 & 342{,}685 & 375{,}295 & 112{,}383 \\
44M  & Per-loop + Raw     & \textbf{42} & \textbf{8} & \textbf{150} & \textbf{390} & \textbf{46} \\
\midrule
129M & Terminal + RMSNorm & 5{,}888 & 5{,}805 & 10{,}636 & 16{,}232 & 1{,}796 \\
129M & Terminal + Raw     & \textbf{21} & \textbf{9} & \textbf{69} & \textbf{305} & \textbf{18} \\
129M & Per-loop + RMSNorm & 26{,}426 & 2{,}132 & 124{,}507 & 149{,}790 & 43{,}792 \\
129M & Per-loop + Raw     & \textbf{71} & \textbf{10} & \textbf{259} & \textbf{808} & \textbf{82} \\
\bottomrule
\end{tabular}
\end{table}

The two RMSNorm-readout failure modes are qualitatively different.
Under terminal-only CE, RMSNorm drift is uniform: median $\approx$ mean and the p99 is within $2\times$ of the median.
Under per-loop CE, RMSNorm drift is heavy-tailed: the median is a few thousand but the mean is $12$--$21\times$ larger and a small fraction of tokens reach $10^5{+}$.
Both raw conditions stay tight at every percentile.

% ============================================================================
\section{Loss-Placement/Scale Decomposition Metrics}
\label{app:endpoint-scale}

Table~\ref{tab:endpoint-scale-full} gives the exact values behind Figure~\ref{fig:decomp}.
The CE gap is $\mathrm{CE}(K{=}1)-\mathrm{CE}(K{=}4)$, so smaller is better for early-exit usability.

\begin{table}[h]
\centering
\small
\caption{Exact loss-placement/scale decomposition values for the controlled 2$\times$2, evaluated with the variable-depth slice harness; per-loop PPL values are the 3-seed means of Table~\ref{tab:depth}, and the 129M terminal values correspond to the cross-entropies in Table~\ref{tab:terminal-depth}. Absolute PPL is therefore lower than the training-time full-validation values in Table~\ref{tab:main}. Per-loop CE marks whether cross-entropy is applied at every loop. Scale-visible marks raw readouts, which expose hidden-state magnitude to the CE loss.}
\label{tab:endpoint-scale-full}
\resizebox{\linewidth}{!}{\begin{tabular}{llccrrrr}
\toprule
\textbf{Model} & \textbf{Condition} & \textbf{Per-loop CE} & \textbf{Scale-visible} & \textbf{$K{=}1$ PPL} & \textbf{$K{=}4$ PPL} & \textbf{CE gap} & \textbf{$r_K$} \\
\midrule
44M & Terminal + RMSNorm & -- & -- & $1.1\times10^{9}$ & 5.04 & 19.21 & $4.1\times10^{-9}$ \\
44M & Terminal + Raw & -- & \checkmark & $6.9\times10^{16}$ & 4.97 & 37.17 & $6.8\times10^{-2}$ \\
44M & Per-loop + RMSNorm & \checkmark & -- & 5.58 & 5.59 & 0.00 & $4.7\times10^{-9}$ \\
44M & Per-loop + Raw & \checkmark & \checkmark & 5.14 & 4.94 & 0.04 & $7.2\times10^{-2}$ \\
\midrule
129M & Terminal + RMSNorm & -- & -- & 250 & 4.66 & 3.98 & $6.9\times10^{-9}$ \\
129M & Terminal + Raw & -- & \checkmark & $4.7\times10^{12}$ & 4.71 & 27.63 & $5.6\times10^{-2}$ \\
129M & Per-loop + RMSNorm & \checkmark & -- & 4.98 & 4.97 & 0.00 & $7.5\times10^{-9}$ \\
129M & Per-loop + Raw & \checkmark & \checkmark & 4.68 & 4.55 & 0.03 & $4.7\times10^{-2}$ \\
\bottomrule
\end{tabular}
}
\end{table}

% ============================================================================
\section{Residual-Update Decomposition}
\label{app:residual-update}

Table~\ref{tab:bounded-block-app} reports the direct measurement of the Lemma~\ref{lemma:expansion} terms on trained checkpoints.
The update magnitude is averaged over loops $k\geq10$, after the initial transient.
The table is not meant to make Lemma~\ref{lemma:expansion} look deep; it checks that the trained checkpoints enter the large-scale regime where the expansion is informative.

\begin{table}[h]
\centering
\small
\caption{\textbf{Normalized-readout checkpoints learn much larger radial residual updates.} Direct measurement of the residual-update decomposition terms from Lemma~\ref{lemma:expansion}.}
\label{tab:bounded-block-app}
\begin{tabular}{llrrrr}
\toprule
\textbf{Model} & \textbf{Condition} & $a_{\rad}$ & $\|b_\perp\|_{\rms}$ & $\|b\|_{\rms}$ & $\|b\|_{\rms}/|a_{\rad}|$ \\
\midrule
44M  & Per-loop + Raw     & $0.27$    & $0.05$  & $0.28$  & $1.01$ \\
44M  & Per-loop + RMSNorm & $257.81$  & $15.45$ & $258.27$ & $1.00$ \\
44M  & Terminal + Raw     & $-0.02$   & $0.06$  & $0.07$  & $3.74$ \\
44M  & Terminal + RMSNorm & $210.81$  & $5.35$  & $210.88$ & $1.00$ \\
129M & Per-loop + Raw     & $0.51$    & $0.08$  & $0.51$  & $1.01$ \\
129M & Per-loop + RMSNorm & $13.99$   & $14.86$ & $20.41$ & $1.46$ \\
\bottomrule
\end{tabular}
\end{table}

% ============================================================================
\section{Additional Mitigation Controls}
\label{app:mitigations}

The main text focuses on the interventions that directly test the readout-blind-spot mechanism.
We also ran mitigation-style controls that help position the result relative to existing looped-transformer designs.

\paragraph{Exit-weighted objective.}
An Ouro-style exit-probability-weighted objective keeps RMSNorm readouts but changes how per-loop CE terms are weighted.
It reduces neither the qualitative drift pattern nor the radial blind spot: final-loop norms remain in the thousands at both scales ($6{,}470 \pm 2{,}233$ at 44M and $4{,}710 \pm 725$ at 129M).
This supports the distinction between changing CE weights and changing what CE can see.

\paragraph{Spectral damping.}
Spectral damping multiplies the hidden state by a learned factor below 1 between loops.
It limits amplification but does not make the readout loss norm-sensitive.
In our runs it lowers norms relative to terminal + RMSNorm ($2{,}432 \pm 238$ at 44M and $2{,}973 \pm 16$ at 129M on the stable seeds), but remains far above raw-readout norms.
One of three 129M seeds became unstable, with hidden-state norms exceeding $10^5$, so we treat it as less robust than exposing scale through the loss.

\paragraph{Inter-loop normalization.}
Ouro uses RMSNorm between loop applications.
This is a different solution: it caps the scale entering each recurrent update rather than making the output loss scale-sensitive.
It is effective, but it changes the loop dynamics and adds normalization inside the recurrent path.

\paragraph{Latent-space losses.}
In denoising-style looped experiments, per-loop latent losses can prevent drift because they directly train $H_k$ in representation space.
This is consistent with our mechanism.
Per-loop losses help scale only when the loss itself is scale-sensitive; CE through a normalized readout is not.

% ============================================================================
\section{Dynamic-Halting Protocol}
\label{app:halt-protocol}

This appendix gives the dynamic-halting protocol summarized in Section~\ref{sec:depth} and Table~\ref{tab:depth}.

\begin{table}[h]
\centering
\small
\caption{\textbf{High halting speedup can reflect $K$-invariance rather than useful adaptive computation.} Calibrated logit-margin halting on per-loop-loss models, mean over 3 seeds, timed slice harness. Thresholds are tuned on a held-out calibration slice to keep dynamic perplexity (PPL) within $1\%$ of fixed $K=4$; the table reports a separate timed slice, so small excesses over the budget reflect calibration generalization.}
\label{tab:halting-main}
\begin{tabular}{llcccc}
\toprule
\textbf{Readout} & \textbf{Model} & \textbf{Fixed $K=4$ PPL} & \textbf{Dynamic PPL} & \textbf{Avg.\ loops} & \textbf{Speedup} \\
\midrule
RMSNorm        & 44M  & 6.01 & 6.01 & \textbf{1.00} & 3.39$\times$ \\
RMSNorm        & 129M & 5.32 & 5.34 & \textbf{1.00} & 3.77$\times$ \\
Raw            & 44M  & 5.29 & 5.34 & 2.16 & 1.55$\times$ \\
Raw            & 129M & 4.86 & 4.91 & 1.76 & 2.09$\times$ \\
Final-only norm & 44M  & 5.37 & 5.41 & 1.78 & 1.90$\times$ \\
Final-only norm & 129M & 4.85 & 4.90 & 1.88 & 1.97$\times$ \\
Norm penalty   & 44M  & 5.36 & 5.41 & 2.60 & 1.29$\times$ \\
Norm penalty   & 129M & 4.85 & 4.90 & 2.56 & 1.43$\times$ \\
\bottomrule
\end{tabular}
\end{table}

\paragraph{Halting decision: per-sequence, not per-token.}
At inference time, the model runs the shared block $K_{\max}$ times.
At each loop $k$, for a sequence of length $L$, we compute a sequence-level confidence score $c_k$ as the mean over tokens of the top-1-vs-top-2 logit margin at that loop:
\begin{equation}
    c_k \;=\; \frac{1}{L}\sum_{t=1}^{L} \bigl(\mathrm{logits}^{(1)}_{k,t} - \mathrm{logits}^{(2)}_{k,t}\bigr),
\end{equation}
where the two logits are the largest and second-largest at position $t$.
A sequence halts at the smallest $k$ for which $c_k$ exceeds a fixed threshold $\tau$.
A sequence that never crosses $\tau$ runs to $K_{\max}$.
We use the mean over tokens rather than the minimum because we found the minimum to be brittle to single-token uncertainty in long sequences---a single hard token at the end forces the whole sequence to spend $K_{\max}$ loops.
Per-token early exit is a different problem (it interacts non-trivially with autoregressive decoding and key-value (KV) caching) and is outside the scope of this paper.

\paragraph{Threshold calibration.}
Given a fixed $K_{\max}{=}4$ and a target relative-PPL budget $\rho{=}1.01$ (within $1\%$ of the fixed-$K_{\max}$ PPL), we sweep $\tau$ on a held-out calibration set of $160$ sequences and pick the smallest $\tau$ such that the resulting dynamic PPL is within $\rho$ of the fixed-$K_{\max}$ PPL on the same sequences.
The calibration is done per checkpoint, so $\tau$ varies across conditions and seeds.
Sweeping $\tau$ uses cached per-loop logits from a single forward pass at $K_{\max}$, so calibration is cheap.

\paragraph{Throughput benchmark.}
After calibration we run a separate timed benchmark on a different 160-sequence slice with the same protocol.
The forward pass uses dynamic batching: at each loop $k$, only sequences that have not yet halted continue to the next loop.
Sequences that halt at loop $k$ leave the batch.
This gives the wall-clock benefit of stopping early.
The reported throughput in the dynamic-halting benchmark is total tokens processed (across all sequences and all positions) divided by total benchmark wall-clock time, including the cost of removing halted sequences from the batch.
Timing setup: single GPU, batch size 8, sequence length 256, bfloat16 (bf16) inference, no KV caching (we evaluate teacher-forced language-model (LM) scoring rather than autoregressive generation).
The fixed-$K_{\max}$ comparison uses the same batch geometry on the same slice.
The reported $\mathrm{Speedup}=\mathrm{tokens/s}_{\mathrm{dynamic}}/\mathrm{tokens/s}_{\mathrm{fixed}\,K_{\max}}$ is therefore mostly attributable to the difference in average loop count, not to batching artifacts.

\paragraph{Caveats.}
This protocol measures speedup in teacher-forced LM scoring, which is the standard variable-depth evaluation in prior work but is not the same as autoregressive generation throughput.
A real-time generation pipeline would interact with KV caching, decoding sampling, and per-step batch-shape changes in ways that we do not measure here.
The protocol is also calibrated on a relatively short held-out slice; with longer sequences or distribution shift, $\tau$ would need recalibration.
We use the same protocol across all conditions, so within-row comparisons are clean even if absolute speedups would differ in a production setting.

\section{Terminal-Only Variable-Depth Baselines at 129M}
\label{app:variable-depth}

Table~\ref{tab:terminal-depth} reports variable-depth cross-entropy for the terminal-only rows of the 2$\times$2 ablation, which we omit from the main-text Table~\ref{tab:depth} because the main variable-depth comparison concerns per-loop-loss exits.
The pattern is informative: terminal + raw has severe degradation at every $K{<}4$ (CE $>13.8$ at $K{=}3$), and terminal + RMSNorm degrades smoothly as depth decreases, from CE $1.54$ at $K{=}4$ to $5.52$ at $K{=}1$.
Both confirm that per-loop CE is necessary for early exits to produce useful predictions, regardless of readout choice.

\begin{table}[h]
\centering
\small
\caption{Variable-depth inference on terminal-only 129M checkpoints (3-seed averages, same variable-depth harness as Table~\ref{tab:depth}), reported as cross-entropy rather than PPL to avoid enormous exponentiated values. Both conditions degrade sharply at $K{<}K_{\max}$, motivating the per-loop-loss focus in the main text.}
\label{tab:terminal-depth}
\begin{tabular}{llcccc}
\toprule
\textbf{Loss} & \textbf{Readout} & $K{=}1$ CE & $K{=}2$ CE & $K{=}3$ CE & $K{=}4$ CE \\
\midrule
Terminal & RMSNorm & 5.52 & 4.80 & 3.18 & 1.54 \\
Terminal & Raw & $>27.6$ & $>16.1$ & $>13.8$ & 1.55 \\
\bottomrule
\end{tabular}
\end{table}

\section{Variable-Depth Analysis of Ouro 1.4B}
\label{app:ouro}

\begin{figure}[t]
\centering
\includegraphics[width=\linewidth]{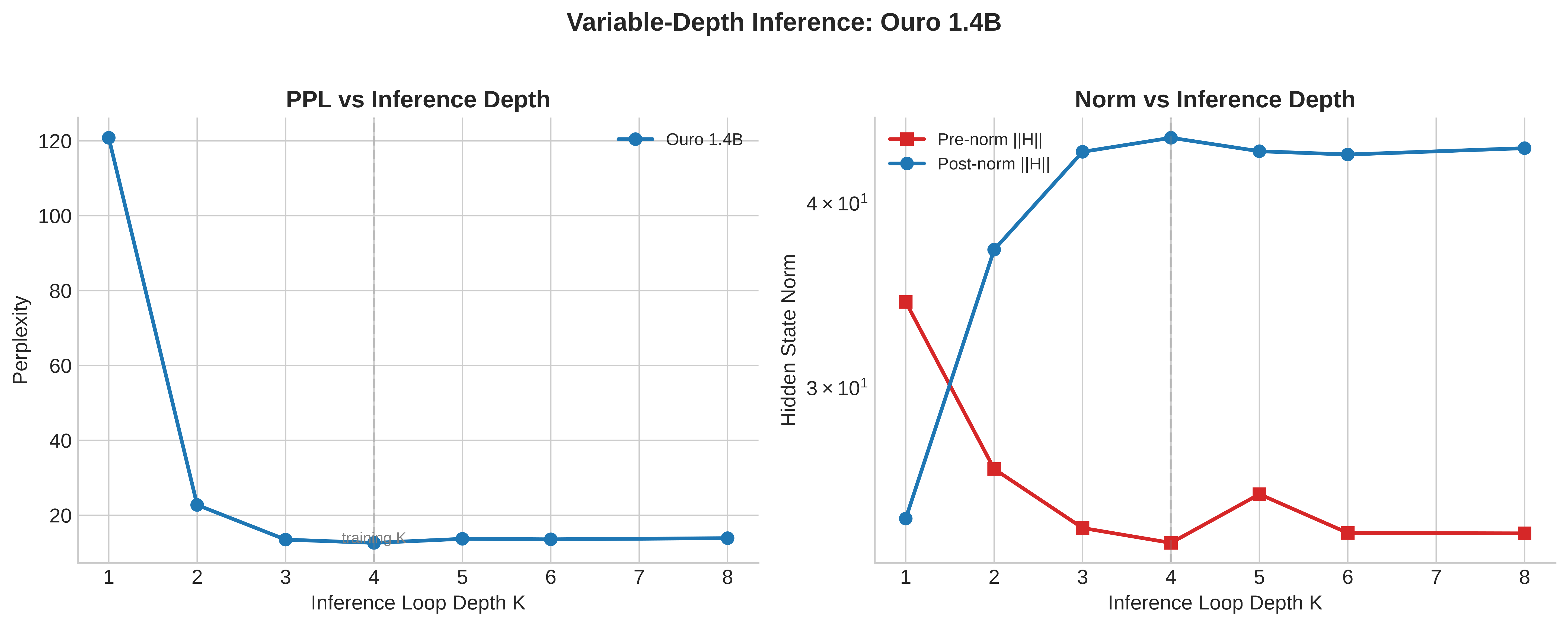}
\caption{Variable-depth inference on the published Ouro 1.4B model. Left: perplexity versus inference depth. Right: hidden-state norms before and after inter-loop RMSNorm.}
\label{fig:ouro_depth}
\end{figure}

We run the published Ouro 1.4B checkpoint \citep{zhu2025ouro} at inference depths $K=1$ through $K=8$ on WikiText-103 validation.
Perplexity converges by $K=3$ and remains essentially flat from $K=4$ to $K=8$ (Figure~\ref{fig:ouro_depth}).
Post-norm hidden-state norms are constant by construction, while pre-norm norms remain low because inter-loop RMSNorm resets scale after every loop.

This result is compatible with our analysis.
Ouro controls recurrent scale by normalizing between loops.
Our per-loop raw and final-only-normalized configurations instead make intermediate CE losses scale-sensitive, allowing the loss to control scale directly.

% ============================================================================

\end{document}